\documentclass[letterpaper, 10 pt, conference]{ieeeconf}  

\IEEEoverridecommandlockouts                              

\usepackage{cite}
\usepackage{amsmath,amssymb,amsfonts}
\usepackage{hyperref}
\usepackage{cleveref}
\usepackage{graphicx}
\usepackage{algorithmic}
\usepackage{graphicx}
\usepackage{textcomp}
\usepackage{colortbl}
\usepackage[table,xcdraw,dvipsnames]{xcolor}

\begin{document}
\title{\LARGE \bf
 AI-Enhanced Automatic Design of Efficient Underwater Gliders
}
\author{
Peter Yichen Chen$^{1\ast}$, Pingchuan Ma$^{1\ast}$, Niklas Hagemann$^{1\ast}$, John Romanishin$^{1}$, \\ Wei Wang$^{2}$,  Daniela Rus$^{1}$, and Wojciech Matusik$^{1}$
\thanks{$^{1}$MIT CSAIL
    {\tt\small \{pyc, pcma, hagemann, johnrom, wojciech, rus\}@csail.mit.edu}}%
\thanks{$^{2}$Marine Robotics Lab, University of Wisconsin-Madison
    {\tt\small wwang745@wisc.edu}}%
\thanks{$^{\ast}$These authors contributed equally to this work.}
}

\maketitle
\thispagestyle{empty}
\pagestyle{empty}

\begin{abstract} 
The development of novel autonomous underwater gliders has been hindered by limited shape diversity, primarily due to the reliance on traditional design tools that depend heavily on manual trial and error. Building an automated design framework is challenging due to the complexities of representing glider shapes and the high computational costs associated with modeling complex solid-fluid interactions. In this work, we introduce an AI-enhanced automated computational framework designed to overcome these limitations by enabling the creation of underwater robots with non-trivial hull shapes. Our approach involves an algorithm that co-optimizes both shape and control signals, utilizing a reduced-order geometry representation and a differentiable neural-network-based fluid surrogate model. This end-to-end design workflow facilitates rapid iteration and evaluation of hydrodynamic performance, leading to the discovery of optimal and complex hull shapes across various control settings. We validate our method through wind tunnel experiments and swimming pool gliding tests, demonstrating that our computationally designed gliders surpass manually designed counterparts in terms of energy efficiency. By addressing challenges in efficient shape representation and neural fluid surrogate models, our work paves the way for the development of highly efficient underwater gliders, with implications for long-range ocean exploration and environmental monitoring. 
\end{abstract}
\section{Introduction}
We introduce an automated design framework for underwater gliders, which are autonomous vehicles optimized for long-range oceanographic sampling due to their low power consumption, efficiency, and lack of external mechanical components \cite{jenkins2003underwater,mahmoudian2008underwater,bender2008analysis,meyer2016glider}. 

Unfortunately, compared to their glider animal counterparts, such as seals and dolphins \cite{khandelwal2023year}, the shapes of existing glider designs lack diversity \cite{matthews2023efficient, an2021parametric}. Although we can partially attribute this lack of diversity to manufacturing constraints, we argue that the primary bottleneck lies in the limitation of design tools. Traditional robot design pipelines rely heavily on manual trial and error: (1) the shape of the object is chosen manually by intuition; (2) the designed shape is then manually fabricated and tested in the real world for its performance. This manual trial-and-error process has limited the capacity to explore more unconventional but potentially more optimized and efficient robot shapes. In the realm of underwater robots, such limitations become even more pronounced. The underwater environment is marked by nonlinear dynamical systems involving complex solid-fluid interactions. Even equipped with analytical tools such as computational fluid dynamics (CFD), each design iteration can take a long time to evaluate. As a result, state-of-the-art underwater robots are still largely characterized by overly simplified shapes.

\begin{figure}[!t]
\centering
\includegraphics[width=1\linewidth]{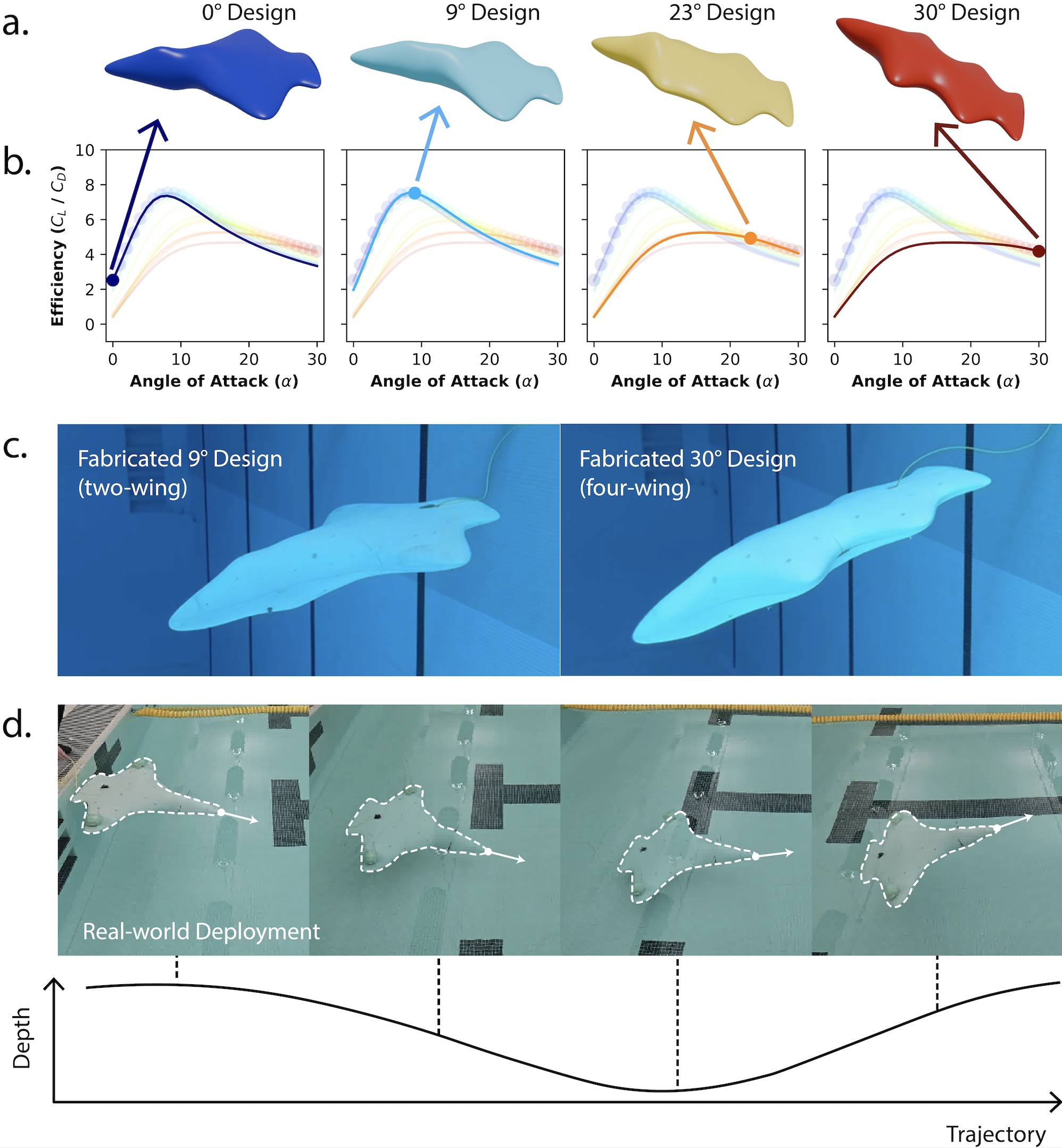}
\caption{\textbf{Efficient underwater gliders.} (a) and (b) Efficient underwater glider designs differ based on the angle of attack, and our algorithm discovers a span of them. We illustrate 4 representative optimal designs for different angles of attack (AoA) and their corresponding efficiency versus AoA curve. (c) Two glider designs were chosen for fabrication and tested as modular outer shells for an internal hardware assembly. (d) We showcase the fabricated designs navigating underwater and the trajectory of the double-wing underwater glider during one dive cycle.}
\label{fig:teaser}
\end{figure}

\begin{figure*}[!t]
\centering
\includegraphics[width=.95\linewidth]{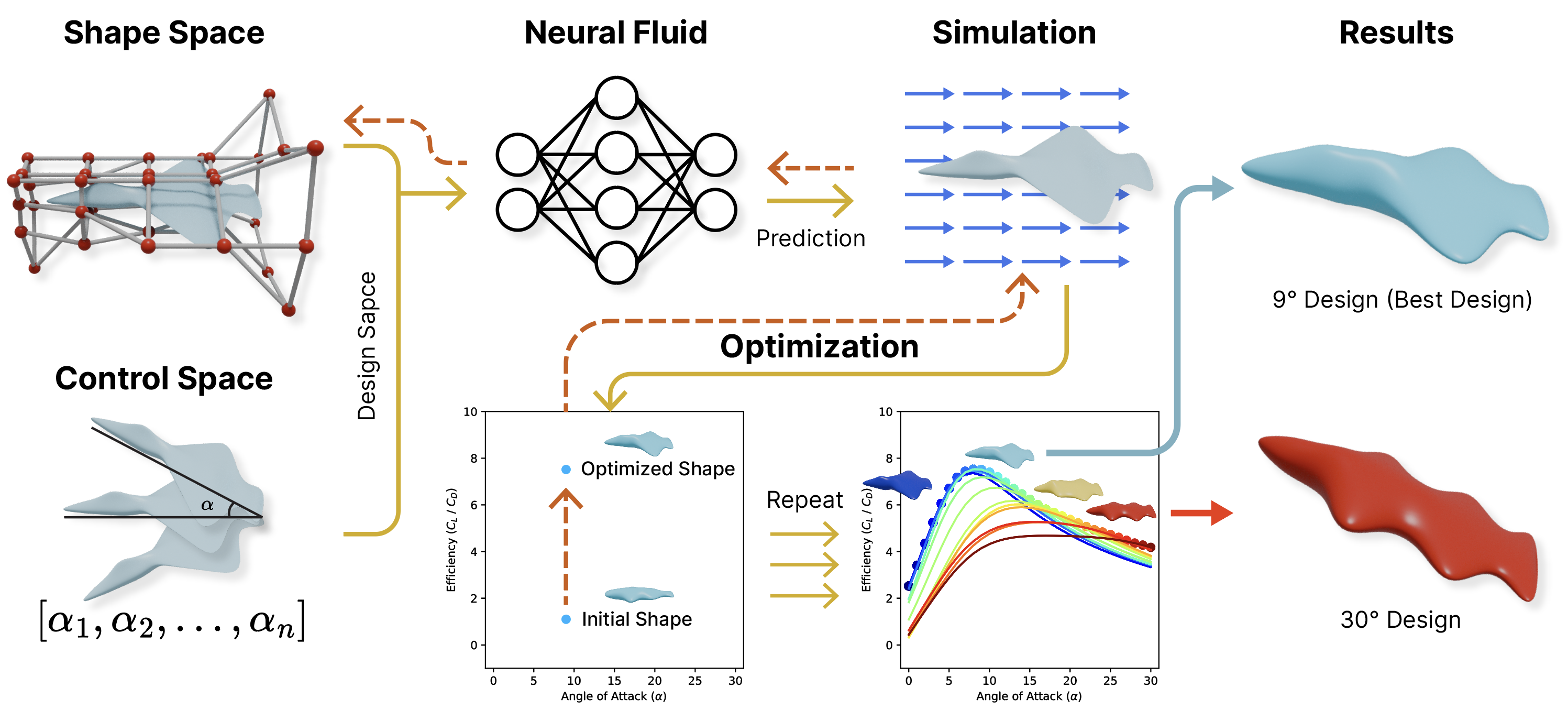}
\caption{\textbf{Computational design framework.} Our co-design framework computes both optimal shape and control for the underwater glider. Shape and control serve as input to our efficient neural fluid model, which efficiently computes the hydrodynamic parameters of the glider. Leveraging these parameters, our framework accurately simulates the performances of these gliders. We then leverage an optimization framework that computationally optimizes their performances (i.e., the lift-to-drag ratio $\eta$), yielding an optimized shape for various angles of attack. On the right, we display two optimal designs for the 9 degrees and 30 degrees.}
\label{fig:computational_design_framework}
\end{figure*}

Here, we propose an alternative, automated approach for designing underwater robots with non-trivial hull shapes \cite{gao2016hull}. Our automated-designed robot outperforms manual-designed robots with traditional shapes in terms of energy efficiency. Although many different approaches to the computational design of underwater robots have been reported in the literature, the automated design of underwater robots to date has been predominantly focused on fine-tuning existing shapes with very few parameters (e.g., Myring Hulls with two parameters) \cite{vardhan2024sample, vardhan2023constrained,won2015design}. Consequently, the variety of existing glider configurations remains significantly constrained, and these designs have not yet approached what could be considered the globally optimal configuration \cite{wang2022development, eriksen2001seaglider, schofield2007slocum}. Representing complex geometries efficiently is impeded by the inherent challenge of balancing expressiveness against computational demands. Furthermore, modeling fluid dynamics adds a layer of complexity. The majority of large-scale fluid dynamics solvers are not directly differentiable \cite{belbute2020combining}, thus precluding their integration into gradient-based optimization workflows, which are pivotal for the concurrent refinement of design and control strategies.


In response to these challenges, we introduce an innovative algorithm designed to transcend these limitations by facilitating the representation of non-trivial geometries and enabling rapid evaluation of their hydrodynamic performance. Central to our approach is the co-optimization of shape and control policies. Specifically, our analysis demonstrates that approaching an optimal, high-efficiency glider design entails achieving the highest possible lift-to-drag ratio. This ratio is, in turn, uniquely determined by both shape and control. Whereas previous glider design work has tended to focus either exclusively on optimizing the shape \cite{yang2021shape,wang2017parametric} or the control \cite{lambert2022experimental}, our work considers both shape and control, thereby enabling significant improvements to be made over archetypal ``torpedo''-shaped glider designs \cite{le2021computational}.

Optimizing both the shape and the control is extremely challenging owing to the complexity of representing varied geometries and the substantial computational expenses associated with computational fluid dynamics. These factors hinder the effective exploration of the design space. In this study, we address these obstacles. Our proposed framework introduces an end-to-end design process spanning all the way from the initial specification of the task to the creation and implementation of an optimally designed robot, including its control strategy. We achieve this by defining the design space of the shapes via a reduced-order geometry representation \cite{xu2021end}. We subsequently construct a differentiable neural-network-based fluid surrogate map that seamlessly maps shape design into lift and drag coefficients. With that, we build an end-to-end differentiable optimization workflow for the automated design of underwater robots. Equipped with this automated design framework, we can discover a wide range of optimal, non-trivial glider designs that are optimal across a range of control settings (\Cref{fig:teaser} and \Cref{fig:computational_design_framework}). Our optimally designed gliders can be directly processed in CAD and allow for fabrication on large-format commercial 3D printers.

Our contributions are fourfold: (1) an efficient and expressive representation of hull shapes; (2) a rapid, differentiable surrogate model for fluid dynamics; (3) simultaneous optimization of shape and control; (4) manufactured designs that surpass existing glider designs in performance.
\section{Method: AI-Enhanced Automatic Design}
\subsection{Preliminaries}
\textbf{Lift-to-drag ratio $\eta = c_l / c_d$} is a critical parameter for determining the efficiency of the glider. Since the glider operates without a propeller, its only energy consumption comes from the buoyancy engine and the mass shifter, with the buoyancy engine being the primary contributor. The buoyancy engine functions by pumping water in or out of the glider, thereby altering its effective density and causing it to either descend or rise. As shown in the appendix, the work required to pump water is directly proportional to $\eta$. Therefore, maximizing $\eta$ by optimizing the glider's shape is essential within our design framework.

\textbf{Angle of Attack (AOA) $\alpha$} is a crucial parameter in controlling the glider, as it significantly affects the hydrodynamic performance of the glider at different values \cite{anderson2011ebook}. Consequently, it is essential for our optimization algorithm to determine the optimal AOA at which the glider should operate to achieve maximum hydrodynamic efficiency.

\subsection{Overview} With these two key parameters in mind, we divide the problem into two specific components: (1) a computational optimization procedure and (2) a real-world hardware realization process. 

In the computational optimization phase, our objective is to design a fabrication-ready shape with optimal hydrodynamic performance. We begin by constructing a design space using a deformation cage representation (\Cref{sec:design_space}), which employs a minimal number of parameters while being capable of expressing all relevant shapes. Next, in \Cref{sec:neural_surrogate}, we develop a neural network-based surrogate model that efficiently computes the fluid dynamics of shapes defined by the deformation cage. This computational phase culminates in solving an optimization problem aimed at maximizing the glider's efficiency ($\eta$) across various angles of attack (AOAs) ($\alpha$).

For the hardware realization phase, we select two shapes from the optimized design set. We then fabricate a modular robotic glider with interchangeable outer shells and conduct field testing in a swimming pool to validate performance.

\begin{figure*}[t]
\centering
\includegraphics[width=.95\linewidth]{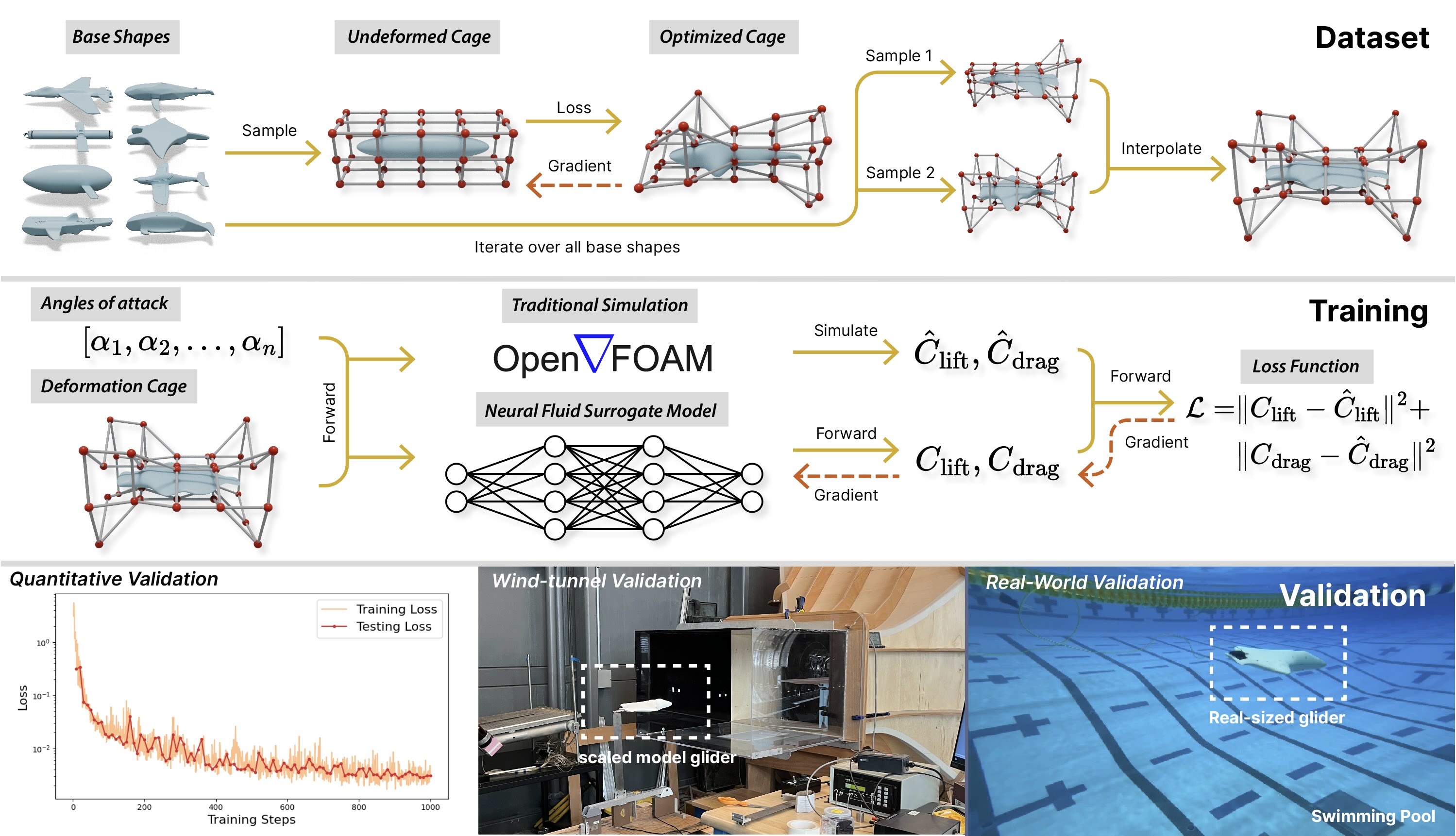}
\caption{\textbf{Efficient computational fluid dynamics.} A key component of our design framework is a neural network-based fluid surrogate model. (Top) Dataset: We begin with a set of base shapes, which are represented using a cage representation. This approach allows for efficient interpolation between shapes. (Middle) Training: Instead of relying on traditional fluid simulators, we train a neural surrogate model to predict drag and lift coefficients based on shape and control parameters. (Bottom) Validation: We validate our neural surrogate model using a testing dataset and further confirm its accuracy through wind tunnel experiments. Finally, we validate the overall design by testing the glider's performance in a swimming pool.}
\label{fig:efficient_fluid_dynamics}
\end{figure*}

\subsection{Efficient Design Space}
\label{sec:design_space}
In co-design problems, shapes are usually parameterized in an expensive manner, using triangle meshes or signed distance fields, to ensure expressiveness. However, these representations come with a large number of parameters. When coupled with complex downstream tasks like underwater fluid dynamics, the efficiency of this kind of approach becomes problematic and prohibits effective optimization. To resolve this issue, the first key element in our computational design framework is a reduced-order shape representation – a 3D Deformation Cage \cite{xu2021end} (see \Cref{fig:computational_design_framework} left and \Cref{fig:efficient_fluid_dynamics} top), where the input parameters are the offsets of the deformation cage's handles and the output of the algorithm is a deformed mesh of the original ellipsoid. We select an ordinary ellipsoid as the initial shape due to its simplicity, generalizability, and excellent initial hydrodynamics properties. Our deformation cage's distinguishing feature is its capacity to represent a wide range of shapes using just very few parameters. As such, in the shape optimization stage, we do not need to optimize with respect to an infinite range, significantly cutting down on computational cost.

To create an expressive shape representation, we manually curated 20 different shapes to serve as foundational structures within our experiments, a subset of which are illustrated in \Cref{fig:efficient_fluid_dynamics}. We include several representative marine animals and man-made aero/hydro-vehicles. We computed deformation cage parameters to approximate these unique shapes via deformations to an initial starting shape. The first step involved generating a sphere and methodically deforming it into an ellipsoid, chosen for its simplistic geometric resemblance to the smooth streamlining of an underwater glider. We subsequently wrapped a deformation cage around this newly formed ellipsoid. This cage acts as a regularization technique in graphics and geometric modeling to maintain the smoothness of transformations~\cite{sorkine2007rigid,jacobson2011bounded}. We applied this procedure across all 20 shapes, achieving smooth transformations and creating a dataset where all geometrical shapes are parameterized identically.

We conducted interpolations between known shapes to enhance our dataset, creating ten morphs for each pair and yielding additional variants. This method expanded our dataset with a comprehensive collection of accurately parameterized and smoothly transformed shapes derived from standard deformable structures. 

\subsection{Neural Fluid Surrogate Model}
\label{sec:neural_surrogate}
We have designed an efficient neural surrogate model to model the interactions between the underwater glider and the surrounding fluids. Our model's input components are the deformation cage parameters and the angle of attack (\Cref{fig:computational_design_framework}), while the outputs are the drag coefficient and lift coefficient. We parameterize our neural fluid surrogate model as a 4-layer multi-layer perception (MLP) with batch normalization. We use $\tanh$ as the nonlinearity in the neural network. In order to improve the data efficiency of the training, we design the neural network to predict the hydrodynamics properties using a single pre-trained model based on different input angles of attack instead of training individual networks for each angle. This augmentation helps us better utilize the statistical correlation within each shape and enlarges our dataset.

To generate the accurate ground truth results necessary for supervision, we utilize OpenFOAM, a standard CFD software package \cite{jasak2009openfoam}. Given any shape represented using triangle meshes, OpenFOAM computes the lift and drag coefficients. In our data generation stage, we set the hyperparameters of OpenFOAM to echo the characteristic values found in sea waters. To expedite the ground-truth data generation process, we employed multithread CPUs. This approach allows us to maximize computational capacity and accelerate the data generation process.

To train our model, we sampled five distinct angles of attack, ranging from -30 degrees to +30 degrees. For the same glider shape, different angles of attack yield varying hydrodynamic performances. In a practical setup, we can control the angle of attack via the mass shifter, which controls the center of mass (CM) of the glider. To fit our neural network surrogate model to the generated data, we utilized the Adam optimizer with a learning rate of 0.1 \cite{kingma2014adam} and a batch size of 128. This process began with a warm-up phase that lasted for 10 epochs. The entire optimization process spanned over 200 epochs. We used MSE loss throughout our experiments. We implemented the neural networks using PyTorch \cite{paszke2019pytorch} and trained all our experiments on a NVIDIA RTX A6000.

\subsection{Optimization Pipeline}
Now that the neural fluid surrogate model has been trained, we will leverage it to optimize the shape of the underwater glider, where the optimization objective is to maximize $\eta$. Given an angle of attack, we initialize the deformation cage by averaging all the shapes that were used for training. Next, we conduct the covariance matrix adaptation evolution strategy (CMA-ES) (\Cref{fig:computational_design_framework} bottom middle) to find the global optima \cite{hansen2006cma}. To ensure all shapes lie within a well-defined range, we constrain this search to fall inside the convex hull defined by the base shapes for the neural fluid surrogate model. We repeat this experiment several times for each angle of attack value. \Cref{fig:teaser}a visualizes the best design for each angle of attack.


\begin{table}[tb]
\centering
\caption{\textbf{Wind-tunnel validation.} The wind tunnel validation yields an average of $4.5\%$ error. The quantity reported is the lift-to-drag ratio.}
\begin{tabular}{|l|l|l|l|l|l|}
\hline
\rowcolor[HTML]{EFEFEF} 
{\color[HTML]{000000} \textbf{AoA}} & {\color[HTML]{000000} \textbf{0°}} & {\color[HTML]{000000} \textbf{4°}} & {\color[HTML]{000000} \textbf{6°}} & {\color[HTML]{000000} \textbf{9°}} & {\color[HTML]{000000} \textbf{Overall}} \\ \hline
\rowcolor[HTML]{FFFFFF} 
Simulation                          & 1.97                               & 5.77                               & 7.05                               & 7.53                               &                                         \\ \hline
\rowcolor[HTML]{FFFFFF} 
Wind Tunnel                         & 1.10                               & 5.35                               & 7.24                               & 7.28                               &                                         \\ \hline
\rowcolor[HTML]{FFFFFF} 
Error                               & +0.87                              & +0.42                              & -0.19                              & +0.25                              & +0.34 (+4.50\%)                         \\ \hline
\end{tabular}
\label{tbl:wind-tunnel}
\vspace{0cm}
\end{table}

\section{Results}
We demonstrate the effectiveness of our computational design framework by automatically designing and fabricating underwater gliders with non-trivial shapes and validating these through real-world experiments.

\subsection{Wind-tunnel Validation}
To validate the accuracy of our neural fluid surrogate model, we 3D-printed the optimal design at the 9-degree angle of attack and performed wind tunnel experiments at the MIT Wright Brothers Wind Tunnel \cite{mitaero2022}. The model is scaled to fit into the wind tunnel (see \Cref{fig:efficient_fluid_dynamics} bottom middle). We ensured Reynold’s numbers matched those of the simulation. As shown in \Cref{tbl:wind-tunnel}, the simulation result and the physical wind tunnel experiments are consistent with each other.

\subsection{Dynamics Simulation}
\begin{figure}[t]
\centering
\vspace{-2em}
\includegraphics[width=\linewidth]{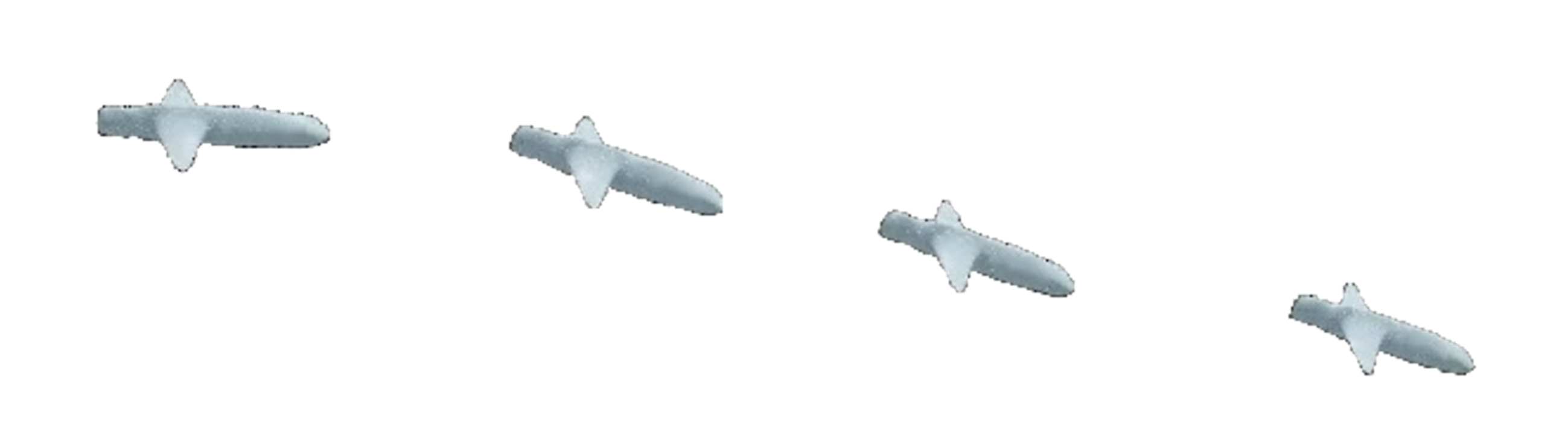}
\caption{\textbf{Dynamics Simulation.} Our dynamics modeling tools accurately simulate the glider's transition from static modes to downward gliding modes.}
\label{fig:dyn_sim}
\end{figure}
In addition to validating our model in the wind tunnel, we also validate the neural surrogate model’s performance in a dynamic modeling setting. The glider was modeled as an articulated body, consisting of the hull shape and a mass shifter connected by a joint to allow for internal movement. We implement this model using Nvdia Warp \cite{warp2022,macklin2024warp} and dynamically time-step through the simulation to observe the glider's motion. The observed gliding motions were primarily a result of the interplay between lift, drag and buoyancy forces, which balanced each other due to the glider's optimal hydrodynamic design, enabling it to glide at a shallow angle. \Cref{fig:dyn_sim} demonstrates the gliding motion. The dynamic simulation provides a predictive visualization of the actual underwater performance.

\subsection{Hardware Design}
To test our optimized glider shapes through real-world underwater experiments, we designed a modular system consisting of a self-contained internal hardware assembly that can be easily re-combined with different outer shells.

\begin{figure}[!t]
\centering
\includegraphics[width=0.9\linewidth]{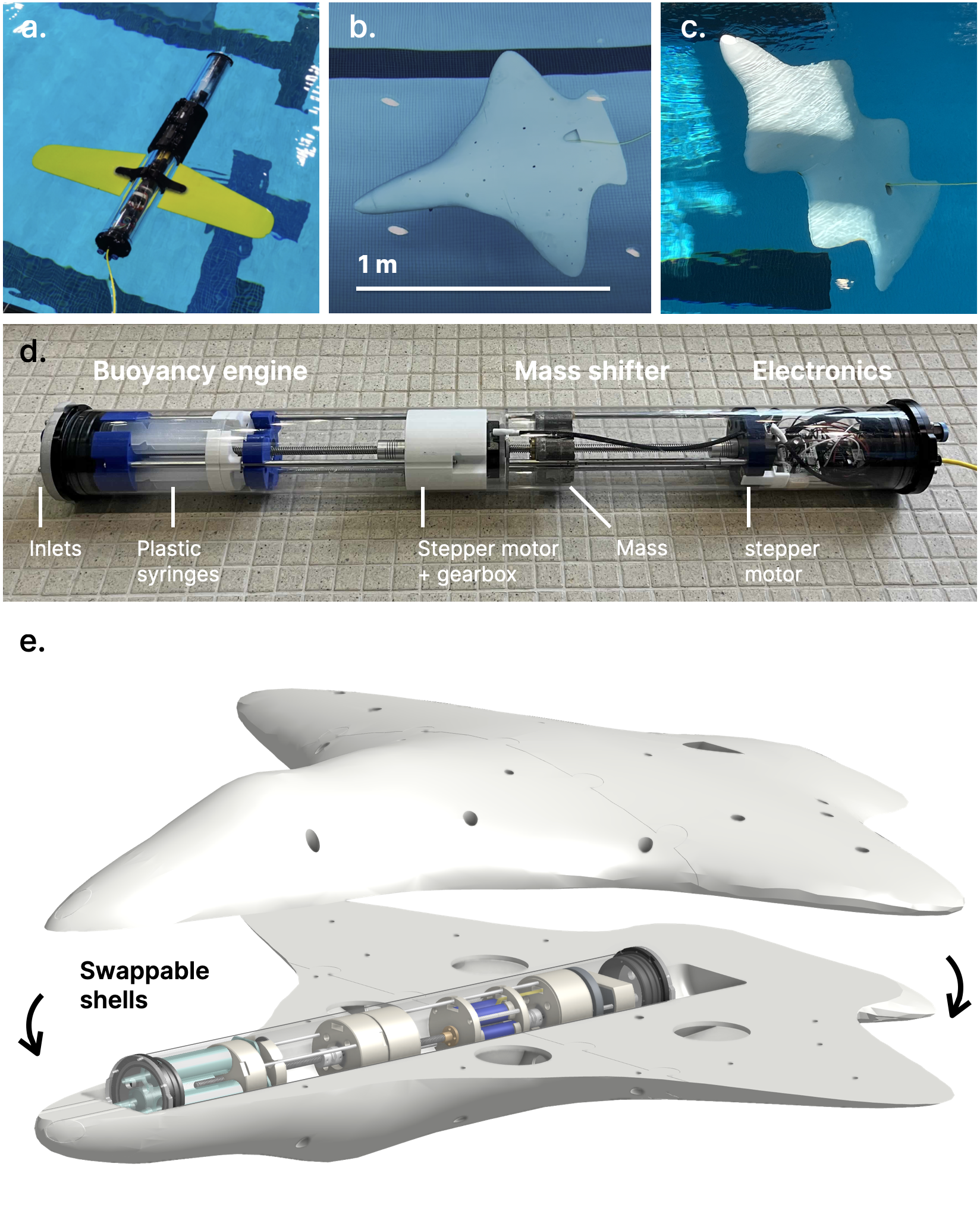}
\vspace{-1em}
\caption{\textbf{Hardware overview.} (a) The internal hardware tube-assembly in a 'traditional' setup with basic wings attached, (b) our optimized two-wing design and (c) our optimized four-wing designs during pool testing. (d) An overview of the internal hardware assembly, including buoyancy engine and mass-shifter. (e) An illustration of how the outer shells assemble with the internal hardware.}
\label{fig:hardware-overview}
\end{figure}

\paragraph{Internal hardware} Similar to other research gliders and open-source designs \cite{osug2017}, the core hardware system of our glider is composed of a buoyancy-engine, mass-shifter, and control electronics assembled within a waterproof tube (see \Cref{fig:hardware-overview}d). The mass-shifter consists of a stepper-motor-actuated lead-screw allowing a mass (e.g. battery pack) of \mbox{0.8 kg} to be translated backward and forward through a stroke of \mbox{16 cm}, shifting the glider's CM and allowing for control over its angle of attack. A small servo motor at the rear allows the mass-shifter assembly to be rotated, enabling roll maneuvers by skewing the CM to one side or the other. The buoyancy engine consists of three plastic plungers actuated by a geared stepper-motor and lead screw: drawing water in to descend, and expelling water to rise. Control electronics are mounted at the rear and consist of a Teensy microcontroller, stepper motor drivers, microSD card for data-logging and 9-DoF inertial measurement unit (IMU) for acceleration and orientation data. The components are mounted on a custom printed circuit board. A pressure sensor mounted at the rear of the tube is also connected to provide depth measurements.

\paragraph{Exchangeable shells}
In designing our glider, we opted for a ``flooded'' design: the shells are hollow and fully flooded with water when submerged, allowing for a light-weight assembly outside of the water. At \mbox{1.1 m} in length, the external shells were broken down into multiple components to fit commercial 3D print beds. The shell components were then assembled into top and bottom halves that could be easily screwed together to form a cohesive structure around the internal hardware tube (see \Cref{fig:hardware-overview}e). When swapping one shell for the other, the internal hardware assembly is left unaltered, streamlining the process of conducting real-world trials. \Cref{fig:hardware-overview}a-c show our two optimized designs next to a more traditional configuration \cite{osug2017}.

\subsection{Underwater Validation}

Using the modular hardware system described above, we conducted underwater experiments for two design configurations generated by our simulation pipeline. We tested a two-wing and four-wing configuration, with optimal efficiencies under an attack angle of 9 degrees and 30 degrees, respectively (\Cref{fig:teaser}). Our two-wing design achieves a horizontal speed of 9.6 cm / s and a vertical speed of 3.8 cm / s. With an effective lift-to-drag ratio of $\eta=2.5$, our optimal design's efficiency significantly outperforms that of the previous standard design in the literature \cite{le2021computational} that reports a lift-to-drag ratio of $\eta=0.3$. This standard design employs a simplistic ``torpedo'' shape and suffers from sub-optimal gliding performance. Our four-wing design achieves a horizontal speed of 7.4 cm / s and a vertical speed of 3.1 cm / s, which translates to a lift-to-drag ratio of $\eta=2.4$, similarly out-performing the basic ``torpedo'' design.


\section{Summary and future work}
In this work, we have developed an end-to-end automated pipeline for designing novel underwater gliders, significantly reducing development time. By expanding beyond the predominant torpedo-like shapes, our approach enables innovative and unconventional glider geometries.

To make good on these promises, several challenges remain to be addressed. One future work direction is in relation to shape representation. Our deformation cage is not ideally suited for representing thin shapes \cite{byun2021underwater}, which may restrict the range of designs that can be explored accurately and could limit applicability to real-world applications. Control, specifically turning and short-distance maneuverability, presents another challenge. As highlighted in \cite{yang2024effective}, achieving precise control in underwater environments is challenging, due to gliders' reliance on buoyancy and internal actuation, which impose limits on their speed and ability to maneuver quickly around obstacles. Similarly, the gliders remain highly susceptible to environmental disturbances like currents and waves, making precise navigation challenging in dynamic underwater environments.

Additionally, we have observed a noticeable simulation-to-reality gap. Our two-wing glider ($\eta=2.5$) does not perform as efficiently in reality as what was modeled in simulation ($\eta=7$). We attribute this gap primarily to frictional forces caused by the surface shear stresses of surface details in the real-world fabricated glider shells. Our CFD modeling pipeline has primarily focused on the overall shapes and does not account for the effects of smaller mechanical details and manufacturing alterations to the original, optimized shape, such as the addition of holes to ensure adequate ``flooding'' and removal of air bubbles during experiments. This oversight may have led to discrepancies between simulated and actual performances. To bridge this gap, future work should consider accurately accounting for these frictional forces. Inspired by \cite{li2024computational}, we also plan to iterate more efficiently between manufacturing and software design processes. This approach will leverage differentiable simulations to not only improve shape design and control strategies but also to enhance the accuracy of gradient optimization.

By addressing the limitations of shape representation and control and by closing the gap between simulation and reality, we can unlock new possibilities in glider design. The potential applications of these gliders are vast, with significant implications for long-range environmental sampling, remote ocean exploration, and marine biology \cite{meyer2016glider}.

\section*{Appendix: Glider Efficiency Analysis}
\begin{figure}[tb]
  \centering
  \includegraphics[width=0.4\textwidth]{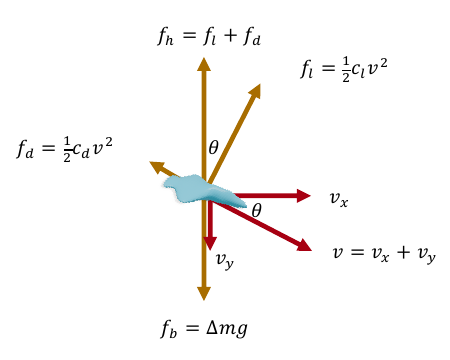}
  \vspace{-2em}
  \caption{\textbf{Force balancing of a buoyancy-engine-driven glider.} Here, $f_d$ is the drag force, $f_l$ is the lift force, and $c_d$ and $c_l$ are the drag and lift coefficients, respectively. $f_b$ represents the total body force, which is the difference between the buoyancy force and the gravitational force. $v$ is the velocity of the glider, comprising both horizontal and vertical components, and $\theta$ is the trajectory angle.}
  \label{fig:efficiency_analysis}
\end{figure}

This section provides a detailed derivation showing that the efficiency of the glider is determined by the non-dimensional lift-to-drag ratio, $\eta = \frac{c_l}{c_d}$. \Cref{fig:efficiency_analysis} illustrates the force balance of a buoyancy-engine-driven glider.

\subsection{Force Balance (Magnitude)}
The magnitude of the force balance can be expressed as follows:
\begin{align*}
    \|f_b\|^2 &= \|f_d\|^2 + \|f_l\|^2,\\
    (\Delta m \cdot g)^2 &= \left(\frac{1}{2} \rho c_d v^2\right)^2 + \left(\frac{1}{2} \rho c_l v^2\right)^2,\\
    (\Delta m \cdot g)^2 &= \left(\frac{1}{2}\right)^2 \rho^2 (c_d^2 + c_l^2) v^4,\\
    \Delta m \cdot g &= \frac{1}{2} \rho (c_d^2 + c_l^2)^{1/2} v^2,\\
    v^2 &= \frac{\Delta m \cdot g}{\left[\frac{1}{2} \rho (c_d^2 + c_l^2)^{1/2}\right]}.
\end{align*}

\subsection{Force Balance (Angle)}
The balance of forces in terms of angle is given by:
\begin{align*}
    \tan(\theta) &= \frac{f_d}{f_l} = \frac{c_d}{c_l},\\
    \cos(\theta) &= \frac{c_l}{(c_d^2 + c_l^2)^{1/2}}.
\end{align*}

\subsection{Combined Analysis}
To analyze the glider's efficiency, we consider the work required to overcome the pressure difference at depth. Assuming the glider operates at a significant depth (e.g., 100 m, approximately 10 atm), with the chamber pressure at 1 atm, the pressure difference ($\Delta P$) can be approximated as:
\begin{align*}
    \Delta P &= \rho g h - P^{\text{chamber}} \approx \rho g h.
\end{align*}
The total work ($W$) performed by the buoyancy engine is computed as:
\begin{align*}
    W &= f \cdot l / \eta^g = \Delta P \cdot A \cdot l / \eta^g = \Delta P \cdot \Delta V / \eta^g\\
    &= \rho g h \cdot \Delta V / \eta^g = g h \cdot \Delta m / \eta^g,
\end{align*}
where $\eta^g$ represents the efficiency of the buoyancy engine. The efficiency in terms of distance covered ($d$) and work done per distance ($W_d^g$) is given by:
\begin{align*}
    d &= 2(h / v_y) \cdot v_x,\\
    W &= 2 \Delta m \cdot g \cdot h / \eta^g,\\
    W_d^g &= \frac{W}{d} = \frac{2 \Delta m \cdot g \cdot h}{2(h / v_y) \cdot v_x} / \eta^g = \frac{\Delta m \cdot g \cdot v_y}{v_x} / \eta^g\\ 
    &= \Delta m \cdot g \cdot \frac{v_y}{v_x} / \eta^g = \Delta m \cdot g \cdot \frac{c_d}{c_l} / \eta^g,
\end{align*}
which shows that the work done per distance is proportional to $\frac{c_d}{c_l} / \eta^g$.
As such, assuming a constant $\eta^g$, the efficiency of the glider is uniquely determined by $\eta=\frac{c_l}{c_d}$. The higher the $\eta$ is, the less work is consumed. From the force balance (angle), we can also see that the smaller the trajectory angle is, the higher the $\eta$ is and the more efficient the glider is.

\section*{ACKNOWLEDGMENT}
This work is supported by Defense Advanced Research Projects Agency (DARPA) grant No. FA8750-20-C-0075 and the GIST-CSAIL research program.


\bibliographystyle{IEEEtran}
\bibliography{reference}






\end{document}